\documentclass[10pt,twocolumn,letterpaper]{article}

\usepackage{iccv}              

\usepackage{url}
\usepackage{multirow}
\usepackage{wrapfig}
\usepackage{colortbl}
\usepackage{bbding}
\usepackage{makecell}
\usepackage{color, xcolor}
\usepackage{float} 
\usepackage{bbding}
\usepackage{booktabs}
\usepackage{graphicx}
\usepackage{caption}
\definecolor{iccvblue}{rgb}{0.21,0.49,0.74}
\usepackage[pagebackref,breaklinks,colorlinks,allcolors=iccvblue]{hyperref}

\def\paperID{1228} 
\def\confName{ICCV}
\def\confYear{2025}

\title{Flow4Agent: Long-form Video Understanding via Motion Prior \\ from Optical Flow}

\author{{Ruyang Liu \footnotemark[1]}
    ~~~{Shangkun Sun \footnotemark[1]}
    ~~~{Haoran Tang}
    ~~~{Ge Li \footnotesize{\Envelope}}
    ~~~{Wei Gao \footnotesize{\Envelope}} \\
    {\small $^1$ School of Electronic and Computer Engineering, Shenzhen Graduate School, ~~~~$^2$Peng Cheng LaboratoryPeking University} \\ 
    {\tt\small \{ruyang@stu., sunshk@stu., hrtang@stu., gaowei262@, geli@\}pku.edu.cn} 
}

\begin{document}
\maketitle

\footnotetext[1]{Equal Contribution}

\begin{abstract}
    Long-form video understanding has always been a challenging problem due to the significant redundancy in both temporal and spatial contents. This challenge is further exacerbated by the limited context length of Multimodal Large Language Models (MLLMs). To address this issue, many previous works have attempted to extract key video information, where the ``key" is typically semantic-aware and heavily dependent on the CLIP model as prior. In this paper, we propose \textbf{Flow4Agent}, a novel framework that pioneeringly incorporates motion priors from optical flow to facilitate LLM-based long video understanding.
    Flow4Agent mitigates the redundancy in long videos at both temporal and spatial levels through two core modules: \textbf{Temporal Granularity Optimization (TGO)} adaptively refines frame-level hierarchies, which first leverages coarse flow priors to group similar visual contents and then applies semantic priors to filter out highly irrelevant scene information. \textbf{Motion Token Pruning (MTP)} further refines the intra-frame visual representations, pruning high-redundancy video tokens using fine-grained optical flow information. Extensive experiments demonstrate that our Flow4Agent outperforms existing methods across a wide range of video MLLM benchmarks, especially for hour-level video understanding tasks, achieving 64.7\% on Video-MME, 71.4\% on MLVU and 60.4\% on LongVideoBench.
\end{abstract}

\section{Introduction}
\label{sec:intro}

\begin{figure*}[t] 
    \centering
    \includegraphics[width=1.0\textwidth]{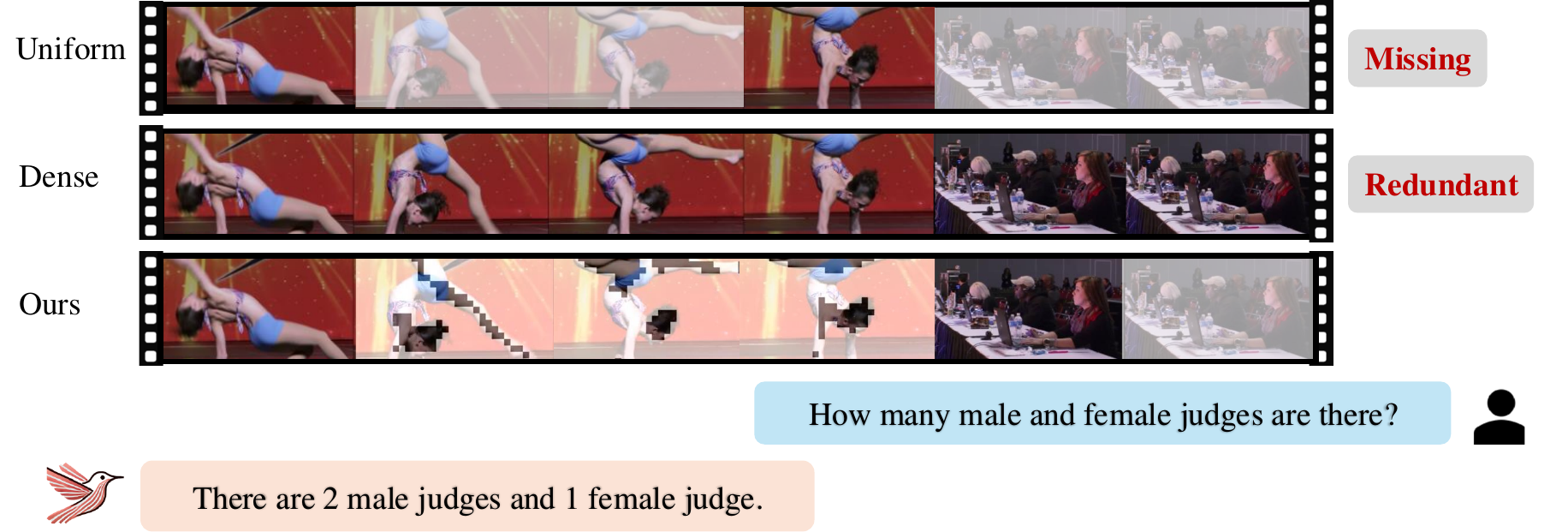} 
    \vspace{-1.5em}
    \caption{Comparison between uniform sampling, dense sampling, and our proposed Flow4Agent.}
    \label{fig:intro}
    \vspace{-0.8em}
\end{figure*}

Multimodal Large Language Models (MLLMs) have made significant strides recently, leading to great changes in various tasks~\cite{vebench, opendmc, wu2024adaptive, chen2023closing}. Thanks to the advancements in LLM \cite{yang2024qwen2, dubey2024llama, hurst2024gpt} and multimodal \cite{wang2024qwen2, llavaonevision, chen2024internvl, liu2024ntire, sun2025ie, sun2025content} pretraining, current MLLMs can effectively interpret visual sequences in images and videos. These models typically support video sequences with hundreds of frames, which is sufficient to cover all contents in short videos, using uniform sampling with whether fixed frame numbers or fixed fps. However, for hour-long videos, this implies that at least one-minute video is distributed to only one frame, leading to substantial information loss, as shown in the first line of Fig. \ref{fig:intro}.

To enable MLLMs to process more video content, one approach is to resample and compress the video tokens \cite{jin2024chat, ppllava, pllava, llamavid}. However, dense resampling inevitably causes the loss of visual information, while frames that can be accommodated by the MLLM remain constrained by a clear upper limit. Another approach is to use memory structures \cite{song2024moviechat, zhang2024flash} or context extension \cite{longva, kangaroo}, enabling the LLM to process densely sampled video frames. However, this method overlooks the widespread information redundancy in long videos. As shown in the second line of Fig. \ref{fig:intro}, the significant redundancy in both time (irrelevant video frames) and space (repetitive content within the same scene) can overwhelm the LLM, resulting in mistakes during long video understanding. 

To address the ubiquitous redundancy in videos, an intuitive solution is to extract key video information. This typically requires additional priors, with the most common being semantic information, such as using the CLIP model to retrieve relevant video content \cite{han2024free, pllava, lvnet, wang2024videoagent, qu2024exploring} or feeding the video's dense captions into the LLM for further reasoning and judgment \cite{wang2024retake, wang2024videotree, wang2024videoagent, yang2025videogen}. However, this reliance on semantic priors has two major drawbacks. First, it heavily depends on the information provided in the user's instructions; when the query offers limited details, much of the effectiveness is lost. Second, the method's performance is constrained by the prior model, such as CLIP models or captioning models, meaning that errors in these models can significantly distort subsequent understanding.

In this paper, we introduce a previously overlooked prior, the motion information from optical flow, to assist in extracting key video content, and propose a novel method, Flow4Agent. Flow4Agent refines the key content in two aspects: inter-frame and intra-frame, which are addressed by the Temporal Granularity Optimization (TGO) module and the Motion Token Pruning (MTP) module, respectively. Specifically, the TGO module utilizes efficient coarse optical flow~\cite{skflow, streamflow} to accurately cluster video scenes, and on this basis, it leverages semantic priors to obtain a sufficiently distinctive set of scenes, thus obtaining highly representative inter-frame features.
Compared to methods that rely solely on semantic priors, our approach is more robust, as we do not depend on semantic priors to directly obtain keyframes. Instead, we use semantic priors to filter out entirely irrelevant scenes, resulting in a lower p-value.
The MTP module, on the other hand, addresses intra-frame redundancy. For highly overlapping video features within the same scene, it uses fine-grained optical flow to filter out the more noteworthy and dynamically varying representations, thereby reducing the overall video redundancy. As the first model to incorporate optical flow for LLM-based video understanding, Flow4Agent neither requires dense captions nor depends on the user-provided details, enabling it to acquire more robust key video content at a lower cost.

To evaluate Flow4Agent, we conducted extensive experiments across a broad range of video understanding benchmarks, including VideoMME \cite{fu2024video}, EgoSchema \cite{mangalam2024egoschema}, Perceptiontest \cite{patraucean2024perception}, MLVU \cite{zhou2024mlvu}, NextQA \cite{xiao2021next}, and LongVideoBench \cite{wu2025longvideobench}. The experimental results demonstrate that Flow4Agent achieves state-of-the-art performance compared to other recent models. Notably, on the three benchmarks with significantly longer videos—VideoMME, MLVU, and LongVideoBench—Flow4Agent attained leading scores of 64.7\%, 71.4\%, and 60.4\%, respectively. Additionally, we evaluated Flow4Agent on various foundational models, further confirming the effectiveness of our motion priors for long-form video comprehension.

The main contributions of our paper are summarized as:
\vspace{-0.3em}
\begin{itemize}
  \item We propose Flow4Agent, which, to the best of our knowledge, is the first model to utilize optical flow information for LLM-based video understanding.
        
 \item We present two novel modules, Temporal Granularity Optimization and Motion Token Pruning, which leverage optical flow from coarse to fine to extract key video content both inter-frame and intra-frame.

 \item We conducted extensive experiments on a wide range of video understanding benchmarks, validating the superior performance of Flow4Agent in video understanding, particularly in long video comprehension.
\end{itemize}

\section{Related Work}
\label{sec:related_work}
\paragraph{Video-based Large Language Models.}
Recent advances in multi-modal large language models (MLLMs) have shown significant progress in processing multi-modal inputs, including video data. Existing research on Video LLMs has focused on both data and model aspects. Regarding data, it has evolved from initially using only video instruction data \cite{videochat,videochatgpt, luo2023valley} to incorporating mixed multi-modal data \cite{llavaonevision, llava-video, apollo, llavainterleave, vila}, which has proven highly effective for video understanding. Additionally, further Reinforcement Learning from Human Feedback (RLHF) after instruction tuning has also demonstrated significant effectiveness for video dialogue \cite{zhang2024direct,ahn2024tuning,cao2024physgame}. At the model level, improvements in the performance of Video LLMs largely depend on advancements in the pretraining of individual components. For example, the more advanced SigLiP~\cite{siglip} surpasses CLIP~\cite{clip} as a visual encoder; models based on the Qwen series~\cite{yang2024qwen2} achieve better results than those using Vicuna~\cite{chiang2023vicuna} as LLM; and Video LLMs built on the latest generation of LLaVA~\cite{llavav15, llavanext, llavaonevision} consistently outperform earlier versions of LLaVA and the BLIP series~\cite{blip2, instructblip}. Meanwhile, compared to the initial mean pooling strategy~\cite{videochatgpt, luo2023valley, videollama} and later approaches incorporating various video-specific components~\cite{btadapter, llamavid, jin2024chat, li2023mvbench}, a more mainstream method has emerged: simply resampling video tokens to directly form a spatiotemporal sequence as input to the LLM~\cite{liu2024st, pllava, llavanext, llava-video, llavaonevision, apollo}. While this paradigm has shown promising results on short videos, the lengthy video sequences and the limited context window of LLMs hinder effective comprehension of long videos.

\vspace{-1em}

\paragraph{Long Video Understanding.}
Fixed-frame sampling remains the predominant choice for most methods. Although some studies have demonstrated the effectiveness of fixed-FPS sampling~\cite{apollo, longvu, llamavid}, the limited context window of LLMs ultimately constrains the number of frames that can be processed. To address this issue, various approaches have been proposed. For example, MovieChat~\cite{song2024moviechat} and Flash-VStream~\cite{zhang2024flash} adopt memory structures and sliding windows to enable streaming video input, while LongVA~\cite{longva} and Kangaroo~\cite{kangaroo} extend the LLM’s context length to accommodate more frames. A more common strategy is video frame resampling, where techniques such as Perceiver-style query sampling~\cite{apollo, llamavid, liu2024st}, clustering~\cite{jin2024chat}, and simple local average pooling~\cite{pllava, ppllava, llavaonevision, llava-video, slowfastllava} all effectively reduce the number of frames. However, long videos inherently contain significant redundancy—temporally, only a few frames carry meaningful information, while spatially, adjacent frames often exhibit high similarity. Thus, feeding the entire long-from video into the LLM is unnecessary.

To this end, a common approach is to extract key content from videos, where the definition of ``key" is guided by additional prior information. For example, LVNet~\cite{lvnet} and LongVU~\cite{longvu} utilize extra visual encoders to compute the similarity between frame features. A more prevalent strategy leverages semantic priors, typically derived from pretrained retrieval and captioning models~\cite{ppllava, longvu, han2024free, wang2024videoagent, wang2024videotree, wang2024retake}. A representative example is VideoAgent~\cite{wang2024videoagent}, which employs both dense captions and CLIP to identify key frames relevant to the user’s query. However, as discussed in Section \ref{sec:intro}, methods relying on semantic priors highly depend on the informative user instruction and accurate prior models. In contrast, our proposed Flow4Agent introduces motion priors from optical flow for the first time, reducing excessive dependence on semantic priors. Additionally, motion information enables a more precise elimination of redundant content.

\section{Approach}
\label{sec:method}

\begin{figure*}[t] 
    \centering
    \includegraphics[width=0.8\textwidth]{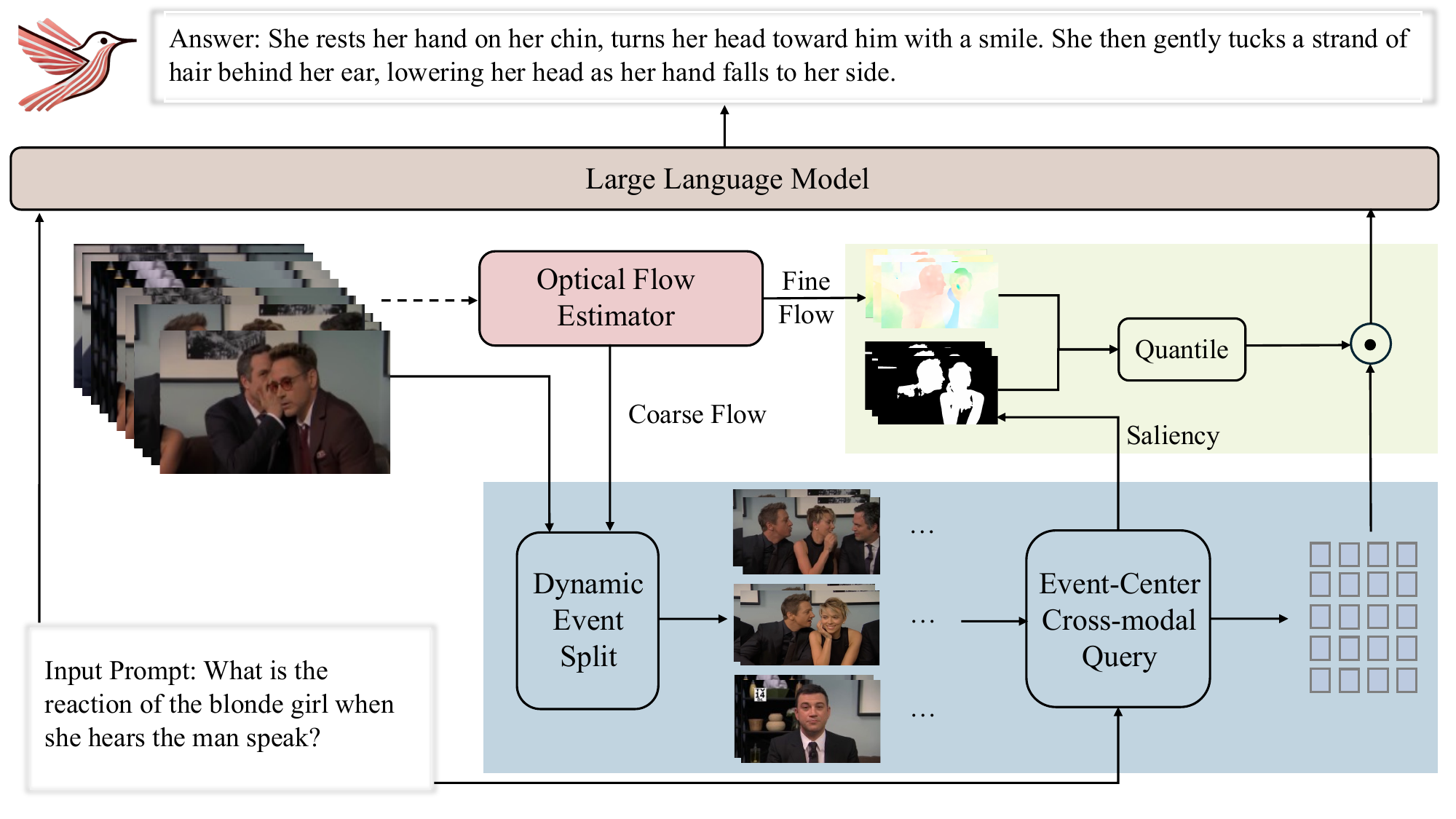} 
    \vspace{-1em}
    \caption{Overview of the proposed Flow4Agent. TGO and MTP strategies are highlighted in the blue and yellow regions, respectively. The dashed line indicates the frames after the first stage in DES, and $\odot$ denotes the Hadamard product operator.}
    \label{fig:overview}
    \vspace{-0.6em}
\end{figure*}

In this section, we will elaborate on how Flow4Agent leverages optical flow to drive LLM-based video understanding. Optical flow has long been an important video understanding prior that provides motion information. In the past, while some previous works have used optical flow for action recognition \cite{simonyan2014two, sun2018optical} or dataset sample filtering \cite{panda}, no research has directly utilized optical flow for LLM-based video understanding. As shown in Fig. \ref{fig:overview}, Flow4Agent addresses redundancy and extracts key content in both inter-frame and intra-frame aspects. In Section \ref{sec:tgo}, the Temporal Granularity Optimization (TGO) module uses HSV transformation with coarse optical flow priors to cluster video content, and employs semantic priors for hypothesis testing to identify representative video content. Then, in Section \ref{sec:mtp}, the Motion Token Pruning (MTP) module uses fine-grained optical flow to capture more significant motion features spatially within high-redundancy events. 

\subsection{Temporal Granularity Optimization} \label{sec:tgo}

Uniform frame sampling, whether using fixed frame numbers or fixed FPS, has been a prevalent approach in previous video understanding models. However, this method often overlooks temporal structural dynamics, leading to limited content diversity. For instance, events with a larger number of frames are likely to be sampled more heavily, even if they lack significant dynamic changes, resulting in redundant frames that contribute little additional information. Conversely, events with fewer frames may be overlooked, even though they might contain crucial information. To address this issue, we propose Temporal Granularity Optimization (TGO) to adaptively refine temporal representation hierarchies in video analysis. As illustrated in Figure~\ref{fig:overview}, the core of TGO lies in a dual-phase spatiotemporal decomposition.

\noindent \textbf{Dynamic Event Split.} We design a motion-aware chromatic analysis strategy to partition a video into different temporal units. The partitioning strategy consists of two stages. In the first stage, we utilize projected temporal differences to divide dynamic events in a coarse manner. Inspired by~\cite{transition}, motion variations of RGB space pixels are often susceptible to factors like illumination changes. Therefore, we first transform frames into the HSV color space, which is less sensitive to illumination variations, thereby eliminating the impact of motion-irrelevant factors like lighting on pixel values. In this space, changes in pixel values better reflect actual event dynamics. 
We then compute the mean square error between consecutive frames, and if it exceeds a threshold, we temporarily mark it as a boundary, completing the coarse first-stage screening.
Given the input video $V$ comprised of frames ${I_1, ..., I_N}$, The first-stage process can be formulated as:
\begin{align}
    V' &= \left\{\Phi(I_{t}) \mid t \in \mathbb{N}\right\}, \\
    \Delta V' &= \left\{\left\|I'_{t+1}-I'_t\right\|_2 \mid t \in \mathbb{N}\right\}, \\
    C &=\{I_t \mid \Delta V'_{t} >\theta, t \in \mathbb{N} \},
\end{align}
where $\Phi$ refers to the HSV transformation and $I'$ denotes the transformed frames. $\theta$ represents the threshold that filters static frames, and $C$ is the set of coarsely selected boundaries in this stage. After that, we leverage pixel-level motion information provided by optical flow to achieve more precise partitioning of dynamic events.  For each potential temporal boundary identified in the first stage, we calculate adjacent $M$ flows within a temporal window. If the maximum magnitudes among these $M$ optical flows exceed a specific threshold, we designate the corresponding frame in the window as the final temporal boundary. In practice, $M$ is set to 3. We employ SeaRAFT~\cite{searaft} for flow calculation, which enables efficient estimation through non-overlapping estimation. Notably, we iterate for only a few rounds to obtain a coarse optical flow, which is sufficient for precise boundary splitting. The entire process can be formulated as:
\begin{align}
    W &= \{ \{I_k\}_{k=t-\left\lfloor\frac{M}{2}\right\rfloor}^{t+\left\lfloor\frac{M}{2}\right\rfloor} \mid I_t \in C \}, \\
    F &= \{\mathcal{E}(w) \mid w \in W \}, \\
    S &= \{I_{\kappa(f_i)} \mid i= \mathop{\arg\max}\limits_{1 \le j \le M} f_{j} > \eta, f \in F\},
\end{align}
where $I_t$ denotes the boundary frame in the first stage with the index of $t$ in the original video. $w$ represents a set of frames in the selected window, and $\mathcal{E}$ refers to the function calculating the corresponding optical flows. $f$ denotes a set of flows in the selected window, and $\eta$ represents the threshold. $\kappa$ is the look-up function that derives the index of the flow in the original video. Traditional key content extraction models often rely on frame-based semantic retrieval or dense captioning. In contrast, once we have divided the video into events, event-centered operations are more efficient and yield better results compared to frame-centered ones, as clustering frames inherently involves the removal of redundant information.

\noindent \textbf{Event-Center Cross-modal Query.}
After dividing the video into events $\{S_i\}$, we select the middle frame of each event to form the anchor frames. This is because within the same event, semantic scenes largely remain unchanged, and thus, frames within the same event are almost identical for an image model. Based on this, we use semantic priors to select keyframes. Previous models typically rely directly on image retrieval models to obtain the most relevant results. Specifically, the target events can be formulated as:
\begin{equation}
    S_{out} = \{S_j| j=\underset{f_i \subseteq S_{i}}{\operatorname{argtopk}} (\Theta_Q(f_i) \cdot \Theta_Q(q))\},
\end{equation}
where $f_i$ is the anchor frame of event $S_i$, $q$ is the embedding of the user’s instruction, and $\Theta_Q$ represents the parameters of image retrieval models such as CLIP or SigLiP.

However, this approach is highly dependent on the accuracy of the prior model $\Theta_Q$. While query-related events may exhibit high similarity, they do not necessarily have the top-k largest similarity scores. Therefore, we introduce two constraints for event selection: first, the selected events should be significant enough to represent the entire video; second, we aim to select as few events as possible to maximize redundancy removal. Therefore, the selected event set $S_{out}$ needs to satisfy the following constraints:
\begin{equation}
\left\{
             \begin{array}{lr}
             \operatorname{min} \operatorname{len}(S_{out}), &  \\
             \alpha(S_i) = \frac{e^{\Theta_Q(f_i) \cdot \Theta_Q(q)}}{\sum_j e^{\Theta_Q(f_j) \cdot \Theta_Q(q)}}, &\\
             \operatorname{p-value} = 1 - \sum_{S_i \subseteq S_{out}} \alpha(S_i) < 0.05, &  
             \end{array}
\right.
\end{equation}
where $\alpha(S_i)$ defines the significance level of the temporal event. When an event contains content strongly relevant to the user's instruction, its significance will outweigh that of other events, and it will be selected independently. Conversely, when the user's instruction lacks sufficient details, these constraints ensure that all important scenes are not overlooked, while filtering out scenes with excessively low significance that are completely irrelevant. Thanks to the integration of motion priors with semantic priors, we are able to adopt a more conservative strategy that filters out a significant amount of redundant content without missing important information.

\subsection{Motion Token Pruning} \label{sec:mtp}



Compared to the numerous works on frame selection strategies for inter-frame redundancy, methods addressing intra-frame redundancy are relatively rare. For example, \cite{ppllava} and \cite{han2024free} use language information to further retrieve key tokens within frames, while \cite{longvu} employs DINOv2~\cite{dinov2} to filter out spatial tokens highly similar to the anchor frame. However, these models also overlook critical motion information. In the same scenes, most of the background remains unchanged, while the small amount of changing foreground information is key to understanding. Therefore, we propose the Motion Token Pruning (MTP) strategy for intra-frame sampling, which utilizes fine-grained motion information to further prune the content within frames.

Specifically, we find that optical flow naturally describes the dynamic information in the scene.
Given a single frame $I_t$, we leverage the pixel-level dense motion information from optical flows to select dynamic-intensive tokens. First, we compute the optical flow between the current and the next frame, which contains not only subject motion information but also other global motions such as camera or background movement. To further eliminate interference from less informative content such as camera motion, we apply homography matrix compensation based on feature points extracted from the predicted flow. Subsequently, we leverage a salient detection mask to obtain the primary motion regions, enabling refined token selection. We then calculate the optical flow magnitude values (after camera motion filtering) for each pixel in these motion regions, select the tokens corresponding to pixels in the top $k\%$ magnitude values, and ultimately generate the final mask to identify valid tokens. In practice, we set $k$ to 50. We utilize U2-Net~\cite{u2net} as the salient detection model, and adopt the powerful Sea-RAFT~\cite{searaft} model to extract accurate optical flows. The entire process can be formulated as,
\begin{align}
f_t &= \mathcal{E}(I_{t}, I_{t+1}), \\
f^*_t &= f_t - \mathcal{H}(I_{t}, I_{t+1}, f_{t}), \\
m_t &= \mathbb{I}(\|f^*_{t}\| \odot s_t \ge Q_{0.5}(\|f^*_{t}\| \odot s_t)), \\
q_t &= p_t \odot m_t,
\end{align}
where $f_t$ refers to the derived optical flow of $I_t$, and $\mathcal{H}$ is the camera motion calculation function based on the homography matrix. $\mathcal{E}$ refers to the optical flow network. $s_t$ denotes the salient detection map of $I_t$ and $\|f^*_{t}\|$ represents the magnitude of the filtered flow. $\odot$ refers to the Hadamard product operator.
$Q_{0.5}$ is the quantile function that select top 50\% pixels with the highest dynamic degree and $\mathbb{I}$ refers to the indicator function. $p_t$ and $q_t$ denote the original token and the filtered token from $I_t$, respectively. Here, we use fine-grained optical flow with more iterations to achieve precise pixel-level pruning.

When segmenting video events using coarse optical flow, we ensure that each unit is assigned at least one frame to prevent information loss. After selecting key events based on semantic priors, we designate these key events along with their neighboring events as priority sampling events, where the number of sampled frames is proportional to the event length. Finally, within each priority-sampled event, we retain all tokens of the anchor frame to preserve complete contextual information, while applying the MTP to further refine intra-frame sampling for the adjacent frames.



\section{Experiment}

\subsection{Implementation Details}
Unless otherwise specified, our experiments are based on the LLaVA-Video-Qwen~\cite{llava-video} extended with Flow4Agent. Results using other basic MLLMs can be found in the ablation studies. Following the standard settings, our image input resolution is 336, the LLM's maximum context length is 8k, and the initial sampling frame count is 64. When performing intra-frame pruning, we simultaneously increase the sampling frame count to maintain the same visual context length as the original model. For motion priors, we use SeaRAFT~\cite{searaft} with 4 iterations in the TGO module and 12 iterations in the MTP module. For semantic priors, we reuse the base model's encoder, SigLiP~\cite{siglip}. All experiments are conducted on two A100 GPUs. Implementation details of each basic model can be found in the appendix.

\subsection{Benchmarks}
We extensively tested the performance of Flow4Agent on six benchmarks, which primarily cover long video understanding and video reasoning. These benchmarks comprehensively evaluate whether our method can identify key video content from highly redundant information. 

\noindent \textbf{VideoMME}~\cite{fu2024video} includes 900 videos of varying lengths and 2,700 manually annotated multiple-choice questions. The video lengths include short ($<$2min), medium (2-30min), and long (30~60min). Since the number of subtitles significantly impacts performance, we adopted a testing setup without subtitles.

\noindent \textbf{LongVideoBench}~\cite{wu2025longvideobench} is a dataset for long video retrieval and reasoning, containing 6,678 manually annotated multiple-choice questions and 17 fine-grained categories. The extraction of subtitles follows the official settings.

\begin{table*}[]
    \centering
\renewcommand{\arraystretch}{1.2}
\setlength{\tabcolsep}{1.5mm}
\caption{Flow4Agent performance on six video benchmarks, including NextQA, EgoChema, PerceptionTest, MLVU, LongVideoBench, and VideoMME. All results are reported as 0-shot accuracy.}
\vspace{-0.5em}
\resizebox{1.\textwidth}{!}{
\begin{tabular}{lcccccccccc}
\toprule  \multirow{2}{*}{\textbf{Models}} & \multirow{2}{*}{\textbf{Size}} & \multirow{2}{*}{ \textbf{NextQA} } & \multirow{2}{*}{ \textbf{EgoSchema} }  & \multirow{2}{*}{\textbf{PercepTest}} & \multirow{2}{*}{\textbf{MLVU}} & \multirow{2}{*}{\textbf{L-VideoBench}} & \multicolumn{2}{p{3.2cm}}{\centering \textbf{VideoMME} } \\ 
&&&&&&& Long & Overall \\
\rowcolor{gray!10} Duration & & 44 sec & 179.8 sec & 16 sec & 3$\sim$120 min &  23sec$\sim$60 min & 30$\sim$60 min & 1$\sim$60 min \\
\midrule
\textit{Proprietary Models} \\
GPT4-V~\citep{Gpt4v} & -  &- & 55.6 & -  & - & 59.1 & 56.9 & 60.7  \\
\midrule
\textit{Open-Source Video MLLMs} \\
Video-LLaVA~\citep{btadapter} & 7B  &- & 38.4 & - & 47.3 & 39.1 & 38.1 & 40.4  \\
LLaMA-VID~\citep{llamavid} & 7B  &- & 38.5 & -  & 33.2 & - & - & - \\
ChatUniVi~\citep{jin2024chat} & 7B &- & - & -  & - & - & 41.8 & 45.9  \\
ShareGPT4Video~\citep{sharegpt4video} & 8B  &- & - & -  & 46.4 & 39.7 & 37.9 & 43.6  \\
LLaVA-NeXT-Video~\citep{llavanext} & 7B & 70.2 & 43.9  & 59.4 & 39.3 & 50.5 & - & 46.5 \\
VideoAgent~\cite{wang2024videoagent} & 7B & 71.3 & 54.1  & - & - & - & - & - \\
VideoTree~\cite{wang2024videotree} & 7B & 75.6 & 61.1  & - & - & - & - & - \\
LVNet~\cite{lvnet} & 7B & 72.9 & 61.1  & - & - & - & - & - \\
VideoLLaMA2~\citep{videollama2} & 7B  & 75.6 & 51.7 & 54.9  & 48.5 & - & 43.8 & 46.6  \\
LongVA~\citep{longva} & 7B  & 69.3 & - & -  & 56.3 & - & 47.6 & 54.3  \\
VideoChat2~\citep{li2023mvbench} & 7B  & - & 54.4 & - & 47.9 & 36.0 & 39.2 & 54.6  \\
LLaVA-OneVision~\citep{llavaonevision} & 7B & 79.4 & 60.1 & 57.1  & 64.7 & 56.4 & 46.7 & 58.2  \\
LLaVA-Video~\citep{llava-video} & 7B & 83.2 & 57.3 & 67.9  & 70.8 & 58.2 & 50.6 & 62.6  \\
Apollo~\citep{apollo} & 7B & - & - &  67.3 & 70.9 & 58.5 & - & 61.3  \\
\rowcolor[RGB]{207,234,241} Flow4Agent & 7B & \textbf{84.0} & \textbf{61.4} & \textbf{69.6} & \textbf{71.4} & \textbf{60.4} & \textbf{54.2} & \textbf{64.7} \\
\bottomrule
\end{tabular}}
\label{tab:main}

\vspace{-0.7em}
\end{table*}

\noindent \textbf{MLVU}~\cite{zhou2024mlvu} is also a benchmark for long video understanding, containing nine different categories with video lengths ranging from 3 minutes to 2 hours, averaging 12 minutes. 

\noindent \textbf{Perception-Test}~\cite{patraucean2024perception} is a benchmark designed to test the perception and reasoning capabilities of MLLMs, containing 11.6k videos and six different annotation types. 

\noindent \textbf{EgoSchema}~\cite{mangalam2024egoschema} contains 5,000 video-question pairs for egocentric evaluation, each video lasting around 3 minutes. 

\noindent \textbf{NextQA}~\cite{xiao2021next} includes 5,440 videos and 49,000 questions, primarily focusing on temporal, causal, and descriptive questions related to video understanding.

\subsection{Main Results}
Table \ref{tab:main} presents a quantitative comparison across multiple video understanding benchmarks. The experimental results show that Flow4Agent consistently outperforms previous state-of-the-art models on all benchmarks.
For instance, compared to the latest model, Apollo, Flow4Agent demonstrates an advantage of 3.9\%, 1.9\%, and 3.4\% on EgoSchema, LongVideoBench, and VideoMME, respectively. Notably, Flow4Agent performs particularly well on long videos. With similar frame sampling and context length, Flow4Agent outperforms LLaVA-Video and LLaVA-OneVision by 3.6\% and 7.5\%, respectively, on VideoMME videos longer than 30 minutes, and surpasses LongVA with a 224k context by 6.6\%. This highlights Flow4Agent’s ability to extract key content from highly redundant videos within a limited context. On benchmarks emphasizing reasoning, such as PerceptionTest and EgoSchema, Flow4Agent also shows strong performance, despite the relatively short length of the videos. This suggests that extracting key information contributes to improving video reasoning capabilities.
Furthermore, compared to other models focused on key video content extraction, such as VideoAgent, VideoTree, and LVNet, Flow4Agent still shows a performance advantage, even though these models are based on the more powerful GPT-4. Interestingly, as a 7B model, Flow4Agent outperforms GPT-4V on most metrics. This demonstrates that motion priors can lead to improvements across different benchmarks.

\subsection{Ablations and Analysis}

\begin{table*}[]
\centering
\setlength{\tabcolsep}{1.5mm}
\caption{Flow4Agent performance on different basic models. We report the results on VideoMME without subtitles. All open-source results are our replication. We applied 4-bit quantization to LLaVA-Video-72B to ensure its deployment on two A100 GPUs.}
\vspace{-0.5em}
\resizebox{0.95\textwidth}{!}{
\begin{tabular}{l|ccc|cccc}
\toprule
\textbf{Model} & \textbf{Context}  & \textbf{LLM Params} & \textbf{Frames} & \textbf{Short} & \textbf{Medium} & \textbf{Long} & \textbf{Overall} \\
  \midrule
LLaVA-NeXT \cite{llavanext} & 4k & 7B & 16 & 54.3 & 41.9 & 38.4 & 44.9  \\ 
\rowcolor{cyan!10} LLaVA-NeXT + Flow4Agent & 4k & 7B & 16 & 55.1 & 44.0 & 42.1 & 47.0  \\ 
LLaVA-OneVision \cite{llavaonevision} & 8k & 7B & 32 & 69.9 & 56.2 & 48.4 & 58.2  \\ 
\rowcolor{cyan!10} LLaVA-OneVision + Flow4Agent & 8k & 7B & 32 & 70.9 & 57.3 & 51.6 & 59.9  \\ 
Qwen2-VL \cite{wang2024qwen2} & 32k & 7B & 64 & 73.0 & 60.8 & 51.3 & 61.7  \\ 
\rowcolor{cyan!10} Qwen2-VL + Flow4Agent & 32k & 7B & 64 & 74.2 & 63.6 & 54 & 63.9  \\
LLaVA-Video \cite{llava-video} & 8k & 7B & 64 & 75.9 & 61.2 & 50.6 & 62.6   \\ 
\rowcolor{cyan!10} LLaVA-Video + Flow4Agent & 8k & 7B & 64 & 77.2 & 62.6 & 54.2 & 64.7 \\ 
LLaVA-Video & 8k & 72B & 64 & 78.0 & 63.7 & 59.6 & 67.1   \\ 
\rowcolor{cyan!10} LLaVA-Video + Flow4Agent & 8k & 72B & 64 & 80.1 & 66.9 & 61.6 & 69.0 \\ 
  \bottomrule
\end{tabular}}
\label{tab:differ}
\vspace{-0.5em}
\end{table*}

\begin{figure*}[t] 
    \centering
    \includegraphics[width=1\textwidth]{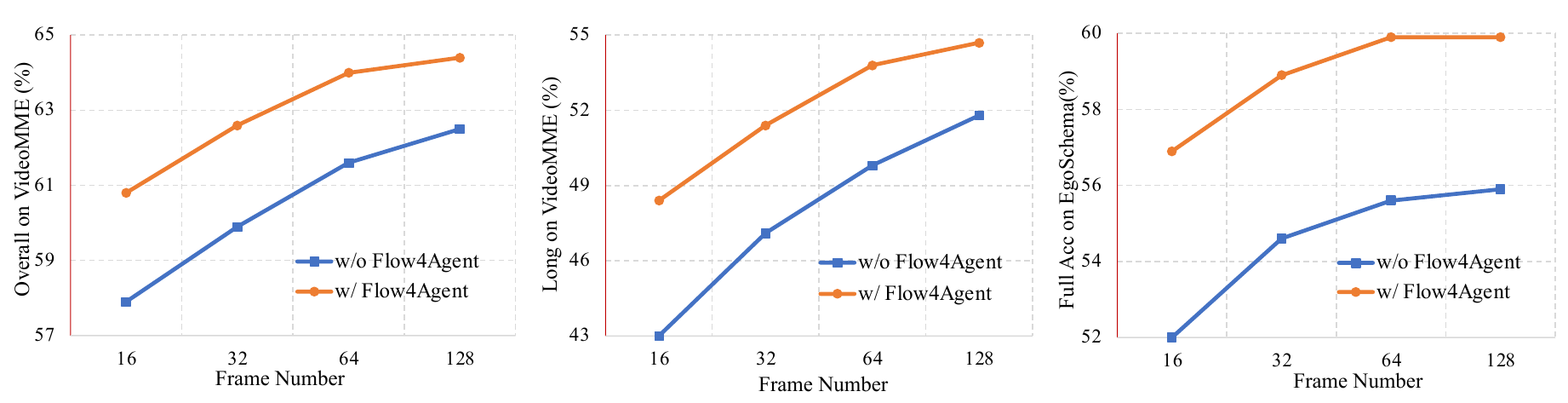} 
    \vspace{-2em}
    \caption{Performance comparison with and without Flow4Agent as the number of frames changes. Flow4Agent provides a greater improvement with fewer frames, while also achieving higher frame efficiency.}
    \label{ablation:frame}
    \vspace{-1em}
  \end{figure*}
  
\begin{table}[]
\centering 
\setlength{\tabcolsep}{0.8mm} 
\caption{The ablation study on model components. DES and ECCQ mean the dynamic event split and event-center cross-modal query respectively in the TGO module.}
\vspace{-0.5em}
\resizebox{0.93\linewidth}{!}{
\begin{tabular}{ccc|ccc|c}
\toprule
\textbf{DES} & \textbf{ECQ} & \textbf{MTP} &  \textbf{Short} & \textbf{Medium} & \textbf{Long} & \textbf{Overall}   \\ \midrule
&  &   & 75.9 & 61.2 & 50.6 & 62.6     \\ 
  \checkmark  &  &   & 77.0 & 61.7 & 50.8 & 63.2    \\ 
    & \checkmark &   & 75.8 & 62.3 & 52.0 & 63.4    \\ 
 \checkmark   & \checkmark &   & 77.1 & 62.2 & 52.9 & 64.0    \\ 
 &  &  \checkmark & 75.9 & 61.5 & 52.4 & 63.3    \\ 
\rowcolor{cyan!10} \checkmark  & \checkmark &  \checkmark & \textbf{77.2} & \textbf{62.6} & \textbf{54.2} & \textbf{64.7}    \\ 
  \bottomrule
\end{tabular}}

\label{abl_comp}
\vspace{-1em}
\end{table}

\noindent \textbf{Performance on Different Base Models.}
To validate the broad effectiveness of Flow4Agent, we conducted integration experiments with various base models. Table \ref{tab:differ} presents the experimental results on VideoMME. We selected different types of models, including pure image models like LLaVA-Next, vision-general models like LLaVA-OneVision and Qwen2-VL, as well as pure video models like LLaVA-Video. The table also includes results with different context lengths, LLM sizes, and video frame counts. It is clearly observed that Flow4Agent consistently provides an improvement across various base models. Moreover, regardless of the underlying model, Flow4Agent shows the most significant improvement for long videos. This highlights the versatility of Flow4Agent as a model-agnostic method, particularly its ability to enhance long-form video understanding.  Additionally, Flow4Agent maintains stable performance across different context lengths, LLM parameter sizes, and sampled frame numbers. 

\noindent \textbf{Effect of Different Components.} 
Flow4Agent consists of two core modules: TGO and MTP, with TGO further divided into motion-guided event splitting and semantic-guided event selection. To assess the impact of these modules, we conducted ablation experiments on the overall model components. In the TGO module, when we use only the motion prior (DES only), frames are allocated to all events directly based on their length. When we use only the semantic prior (ECQ only), semantic checks are performed on individual frames rather than events. As shown in Table \ref{abl_comp}, the two components within TGO provide significant gains for short and long videos, respectively, and their combination leads to improved performance across all video lengths. The MTP module further enhances long video understanding. Each module complements the others, collectively demonstrating the design of Flow4Agent.

\begin{figure*}[ht] 
    \centering
    \includegraphics[width=1.0\textwidth]{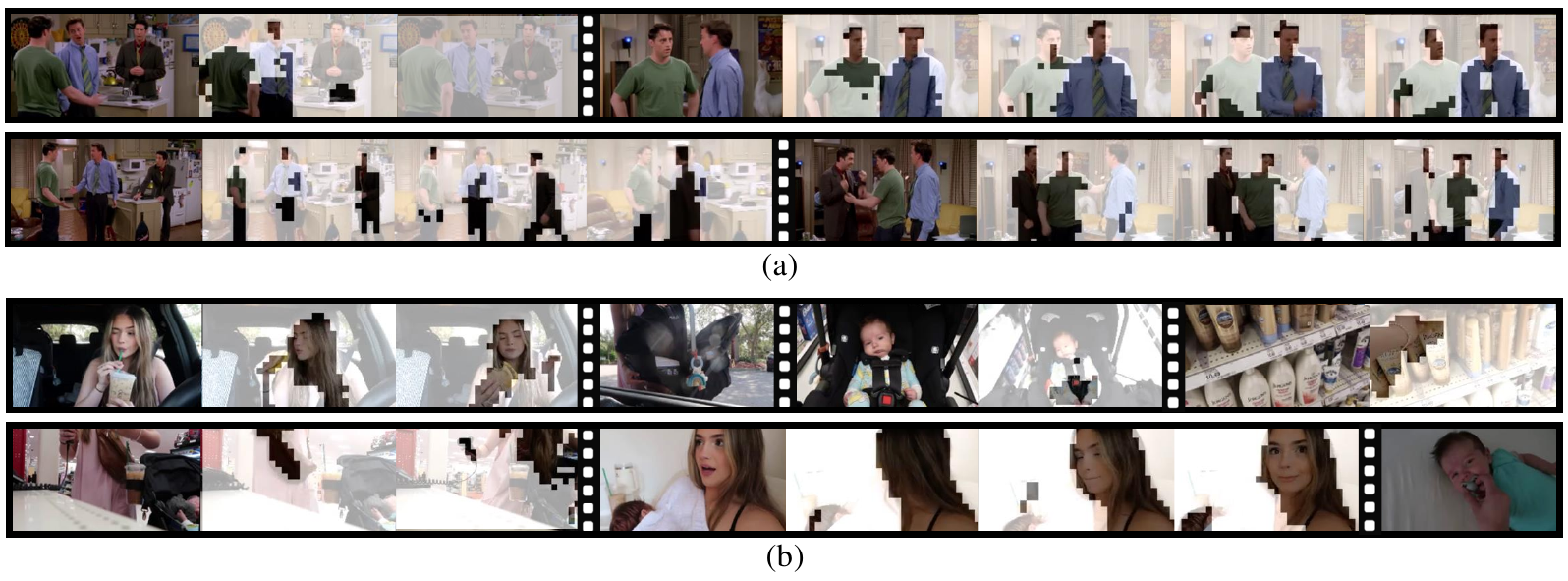} 
    \vspace{-2em}
    \caption{Visualizations of how Flow4Agent reduces redundancy.}
    \label{fig:demo}
    \vspace{-0.8em}
\end{figure*}

\noindent \textbf{Effect of Different Frames Number.}
The number of frames is a crucial variable affecting video understanding, particularly for long videos. In theory, a sufficient number of frames ensures comprehensive coverage of necessary information to answer a given question. However, an excessive number of frames can introduce redundant or irrelevant information, potentially overwhelming the model. Thus, the ability to extract key information within a constrained frame budget is a critical metric for evaluating model performance. As shown in Fig. \ref{ablation:frame}, we analyzed performance variations across different frame counts using three test sets with varying video lengths: VideoMME-Overall, VideoMME-Long, and EgoSchema. To support a 128-frame input, we adjusted the avgpooling2d kernel size from 2 to 3. The results indicate that Flow4Agent consistently enhances performance regardless of the number of input frames. When frame availability is limited, Flow4Agent’s advantage becomes even more pronounced. Additionally, Flow4Agent achieves performance saturation with fewer frames, demonstrating higher frame efficiency. These findings highlight Flow4Agent’s effectiveness in extracting critical video information while optimizing frame utilization.

  \begin{table}[]
\centering 
\setlength{\tabcolsep}{0.8mm} 
\caption{The ablation study on the motion-prior model. Iter-TGO and Iter-MTP refer to the number of iterations of the optical flow model within the TGO and MTP modules, respectively.}
\vspace{-0.5em}
\resizebox{0.9\linewidth}{!}{
\begin{tabular}{l|cc|cc}
\toprule
\textbf{Flow Model} & \textbf{Iter-TGO} & \textbf{Iter-MTP} & \textbf{Long} & \textbf{Overall}   \\ \midrule
NeuFlow & 4 & 12   & 53.0 & 64.1     \\ 
StreamFlow & 4 & 12   & 53.9 & 64.5     \\ 
\rowcolor{cyan!10} Sea-RAFT & 4 & 12   & 54.2 & 64.7     \\ 
Sea-RAFT & 12 & 12   & 54.4 & 64.6     \\ 
Sea-RAFT & 4 & 4   & 53.3 & 64.2     \\ 

  \bottomrule
\end{tabular}}

\label{abl_flow}
\vspace{-0.6em}
\end{table}

  \begin{table}[]
\centering 
\setlength{\tabcolsep}{0.8mm} 
\caption{The ablation study on semantic-prior model.}
\vspace{-0.5em}
\resizebox{0.9\linewidth}{!}{
\begin{tabular}{l|cc|ccc}
\toprule
\textbf{CLIP Model} & \textbf{Size} & \textbf{Resolution} &  \textbf{Long} & \textbf{Overall}   \\ \midrule
OpenAI-CLIP  & 0.4B & 224  & 53.3 & 63.9     \\ 
EVA-CLIP & 8B & 224  & 53.1 & 64.0     \\ 
SigLIP & 0.4B & 224   & 53.9 & 64.2     \\ 
\rowcolor{cyan!10} SigLIP & 0.4B  & 336   & 54.2 & 64.7     \\ 
 
  \bottomrule
\end{tabular}}

\label{abl_clip}
\vspace{-0.9em}
\end{table}

\noindent \textbf{Effect of the Prior Model.}
Motion and semantics are two essential priors leveraged by Flow4Agent, and the choice of corresponding models as well as their configurations can significantly influence the final performance. In Table \ref{abl_flow}, we examine different optical flow models and their iteration counts within the TGO and MTP modules. While more iterations yield finer optical flow information, they also increase computational time. Our results show that the state-of-the-art Sea-RAFT model delivers superior performance. Additionally, in the TGO module, fewer iterations can obtain optimal results, whereas the MTP module benefits from more iterations for the best performance. This highlights the optimal configuration of Flow4Agent: coarse optical flow for event splitting and fine-grained optical flow for visual token pruning. In Table \ref{abl_clip}, we explore various semantic prior models. The SigLIP-336 model achieves the best results, demonstrating that stronger semantic priors contribute to improved performance.

\subsection{Visualization}
To qualitatively assess the effectiveness of Flow4Agent, we present several visualization cases in Fig.~\ref{fig:demo}. Different scenes identified by the TGO module within the same video are separated by film lines, while redundant regions filtered out by the MTP module within each scene are grayed out. Across all cases, we observe that the TGO module effectively differentiates distinct scenes. For instance, in Fig.~\ref{fig:demo}(a), despite the highly similar background, the TGO module successfully distinguishes between a two-person conversation and a three-person group interaction. Additionally, the MTP module efficiently removes redundant background elements within the same scene while preserving crucial variations, such as human actions and facial expressions. More examples can be found in the appendix.

\vspace{-0.3em}

\section{Conclusion}
In this paper, we propose Flow4Agent, which introduces optical flow into LLM-based video understanding for the first time, incorporating a novel motion prior to extract key video information. Flow4Agent enhances long-form video comprehension through two modules: the Temporal Granularity Optimization (TGO) module that leverages coarse motion and semantic information to eliminate inter-frame redundancy and identify key events, and the Motion Token Pruning (MTP) module that utilizes fine-grained optical flow to remove intra-frame redundancy. Extensive quantitative and ablation experiments demonstrate the effectiveness of Flow4Agent in long-form video understanding and video reasoning, achieving state-of-the-art performance across a wide range of video benchmarks.

{\footnotesize{
\noindent \textbf{Acknowledgements.} 
    This work was supported by National Science and Technology Major Project (2024ZD01NL00101).}}

{
    \small
    \bibliographystyle{ieeenat_fullname}
    \bibliography{main}
}

\clearpage
\appendix
\section{More Implementation Details}
\noindent \textbf{Pooling Stragety.}
Our pooling strategy remains consistent with the approach used during the pretraining of the base models. For LLaVA-Next, we do not apply any pooling. For LLaVA-OneVision and LLaVA-Video-72B, we use bilinear interpolation to reduce the width and height of each frame to half of their original size. In contrast, for Qwen2-VL and LLaVA-Video-7B, we directly apply average pooling with a kernel size of 2 (avgpooling-2d). For these four models, the final number of tokens per frame input is $13*13=169$.

\vspace{0.2em}

\noindent \textbf{User Prompt.}
For all models, the system prompt is uniformly set as ``You are a helpful assistant." In LLaVA-Next and Qwen2-VL, we did not add an additional question prompt. For multiple-choice QA, in LLaVA-OneVision, the question prompt before the question is set as ``Select the best answer to the following multiple-choice question based on the video and the subtitles. Respond with only the letter (A, B, C, or D) of the correct option.", while the prompt after the question is ``Answer with the option's letter from the given choices directly." In LLaVA-Video, the question prompt before the question is set as ``The input consists of a sequence of key frames from a video. Answer the question concisely first and followed by significant events, characters, or objects that appear throughout the frames. Question:", and the prompt after the question is ``The best answer is:". For open-ended QA, we did not set any additional question prompt.

\vspace{0.2em}

\noindent \textbf{Evaluation Benchmark Details.}
For all benchmarks, the reported metric is accuracy, which is the ratio of correctly answered questions to the total number of questions. The results for EgoSchema and PerceptionTest are obtained from the official evaluation server, while the results for VideoMME, MLVU, NextQA, and LongVideoBench are computed locally. For VideoMME, we did not use any subtitles because we found that different models use varying amounts of subtitles—most models only use subtitles corresponding to the sampled frames, whereas LLaVA-Video and LLaVA-OneVision use nearly all available subtitles. For LongVideoBench, to ensure a fair comparison, we used the subtitles corresponding to the sampled frames and placed all utilized subtitles together before the question.

\section{More Visualizaion}
In the supplementary materials, we present several tri-frame GIFs illustrating the video, its optical flow, and the final mask. As shown, optical flow serves as an effective tool for extracting motion priors, successfully identifying most of the informative regions within a video. Whether in relatively static scenes (e.g., people engaged in conversation) or highly dynamic scenarios (e.g., a track and field race), our method accurately captures key subjects while maintaining high consistency and stability within the same scene or event. This extraction of critical information subsequently helps improve long-form video understanding in MLLM, even with a limited number of frames. Additionally, since we retain a complete video frame for each scene, we ensure that no crucial information is lost during the extraction process. These observations highlight the effectiveness of Flow4Agent in capturing and focusing on essential video content, even in complex scenes with dynamic backgrounds.

\section{Limitation}
Our research is primarily concentrated on video understanding tasks, particularly in the context of long videos. As a result, Flow4Agent provides limited improvements for short videos and images. For short videos, uniform frame sampling can almost fully capture the video content, making additional key information extraction less beneficial. Moreover, optical flow cannot be applied to images or disjointed multi-image inputs, leading to no improvement in image-based tasks. Additionally, training could potentially bring further enhancements, but due to the constraints in GPU resource, we leave it as a future work to explore instruction tuning combined with Flow4Agent.

\end{document}